\pdfoutput=1

\documentclass[11pt]{article}

\usepackage[preprint]{acl}

\usepackage{times}
\usepackage{latexsym}

\usepackage[T1]{fontenc}

\usepackage[utf8]{inputenc}

\usepackage{microtype}

\usepackage{inconsolata}

\usepackage{graphicx}
\usepackage{amsmath} 

\title{Intrinsic Bias is Predicted by Pretraining Data and Correlates with Downstream Performance in Vision-Language Encoders}

 \author{Kshitish Ghate\thanks{Equal contribution.} \\
   Carnegie Mellon University \\
  \texttt{kghate@cs.cmu.edu} \\\And
   Isaac Slaughter\footnotemark[1]  \\
   University of Washington \\
  \texttt{is28@uw.edu} \\\And
  Kyra Wilson \\
   University of Washington \\
  \texttt{kywi@uw.edu} \AND
  Mona Diab\\
   Carnegie Mellon University\\
  \texttt{mdiab@andrew.cmu.edu} \\\And
  Aylin Caliskan\\
   University of Washington\\
  \texttt{aylin@uw.edu}
}

\begin{document}
\maketitle
\begin{abstract}
While recent work has found that vision-language models trained under the Contrastive Language Image Pre-training (CLIP) framework contain intrinsic social biases, the extent to which different upstream pre-training features of the framework relate to these biases, and hence how intrinsic bias and downstream performance are connected has been unclear. In this work, we present the largest comprehensive analysis to-date of how the upstream pre-training factors and downstream performance of CLIP models relate to their intrinsic biases. Studying 131 unique CLIP models, trained on 26 datasets, using 55 architectures, and in a variety of sizes, we evaluate bias in each model using 26 well-established unimodal and cross-modal principled Embedding Association Tests. We find that the choice of pre-training dataset is the most significant upstream predictor of bias, whereas architectural variations have minimal impact. Additionally, datasets curated using sophisticated filtering techniques aimed at enhancing downstream model performance tend to be associated with higher levels of intrinsic bias. Finally, we observe that intrinsic bias is often significantly correlated with downstream performance  ($0.3 \leq r \leq 0.8$), suggesting that models optimized for performance inadvertently learn to amplify representational biases. Comparisons between unimodal and cross-modal association tests reveal that social group bias depends heavily on the modality. Our findings imply that more sophisticated strategies are needed to address intrinsic model bias for vision-language models across the entire model development pipeline. \textcolor{red}{Warning: This study contains figures and information that may be triggering and/or offensive to readers.}

\end{abstract}

\section{Introduction}
Neural network models are prone to learning patterns based on statistical associations between concepts within their training data that might lead to harmful bias when it relates to social groups or model performance \citep{fabbrizzi2022survey}. This phenomenon has been observed in a number of vision and language models, each of which are unimodal and learn information within a single modality \citep{Caliskan2017SemanticsBiases, Guo2021DetectingBiases, Steed2021, Wolfe2022VAST:Models, omrani2023evaluating}. Cross-modal models, which learn information from both vision and language modalities, also learn biased information relating to social group associations \cite{Goh2021MultimodalNetworks,Wolfe2022AmericanAI,  Wolfe2022EvidenceAI,
Wolfe2022MarkednessAI, Wolfe2023ContrastiveBias, Janghorbani2023Multi-ModalModels, Berg2022ALearning, mandal_multimodal_2023, hall_visogender_2023}.

These results have largely been found using intrinsic bias tests adapted from Natural Language Processing (NLP): evaluations that compare relative distances between a model's representations of stimuli representing different concepts and social groups. Despite their prevalence in model evaluation, however, there is limited work connecting them to other factors of model design and optimization. For example,  upstream factors such as training datasets, model architectures, and model sizes directly determine the representations learned and consequently may be reflected in intrinsic bias tests. These representations are then directly used for downstream tasks such as zero-shot image classification, which suggests a potential connection between the intrinsic bias of a model and its performance on downstream tasks. By investigating intrinsic bias as it explicitly relates to these two upstream pre-training and downstream zero-shot performance factors, we are able to draw novel insights about the ways in which models can be optimized to reduce harmful or undesirable biases.

We measure the associations between social groups and valence, the pleasantness or unpleasantness of a concept \citep{toney2020valnorm}. Valence is a robust dimension of human cognition as it relates to shaping attitudes and biases \citep{harmon2013does}. Social groups and valence associations also exist in unimodal models \citep{ Wolfe2022VAST:Models}. In this work, we examine whether these associations are observed cross-modally and how they relate to upstream factors and downstream model performance. 

To our knowledge, our work is the first which connects 26 tests of intrinsic bias, including those related to race, gender, age, and baseline associations with respect to non-social group concepts such as flowers and insects, to upstream factors of model training (including 26 training datasets, 55 model architectures, and model sizes ranging between 100 million and 5 billion parameters). While variations on all of these features have been proposed to either mitigate biases or improve performance on downstream tasks, we are the first to address the variance in the magnitude of the effect of these features in CLIP models. Additionally, we connect intrinsic bias tests to a suite of 35 zero-shot image classification and retrieval tasks \cite{Schuhmann2022LAION-5B:Models}. A novel contribution of our work is that we show that optimizing for performance is not sufficient to mitigate intrinsic biases. The scale of our experiments allows us to obtain high statistical power in analyzing the relationship between intrinsic bias and both upstream factors and downstream performance and making the following generalizable knowledge contributions about cross-modal models\footnote{We release our code and data at \url{https://github.com/kshitishghate/CLIP\_bias/}.}:

\textbf{1}. By improving the application of EATs via controlling the valence of images and text used in EATs, we decreased variance in effect size of bias on average by 4.8\% across unimodal and cross-modal tests, and demonstrated significant intrinsic bias in 131 models across 26 EATs. Aggregate intrinsic bias is consistent with human associations in 78.86\% of the 3,406 cases and varies by modality combination and test categories. 
    
\textbf{2}. We demonstrate that the choice of training dataset significantly impacts intrinsic bias, independent of other upstream factors such as model architecture or parameter count. Notably, while current dataset filtering techniques \cite{gadre2024datacomp, fang_data_2023, xu_demystifying_2023} have been successful in optimizing performance metrics like ImageNet classification accuracy, they fall short in addressing fairness. Moreover, filtering methods driven by automated neural network decisions \cite{fang_data_2023}, despite yielding better downstream results compared to heuristic-based approaches \cite{gadre2024datacomp}, tend to exacerbate societal biases even further (e.g. with a $\beta=0.608$ over baselines). Our findings provide strong evidence that bias amplification often originates from choices made during the upstream data curation process.

\textbf{3}. We show that intrinsic bias measures are correlated with downstream performance. Across modality settings, higher intrinsic bias often correlates with improved performance for non-human associations as seen with `Flower-Insect/Valence' (aggregate $r=0.56$) and `Instrument-Weapon/Valence' (aggregate $r=0.78$), suggesting consistent training signals may amplify certain associations. For `Gender/Valence' tests ($r=-0.51$, $r=-0.27$ in two modality settings), improved performance increased positive associations for men, indicating non-congruent emergent stereotypes.

\begin{figure}
    \centering
    \includegraphics[width=\linewidth]{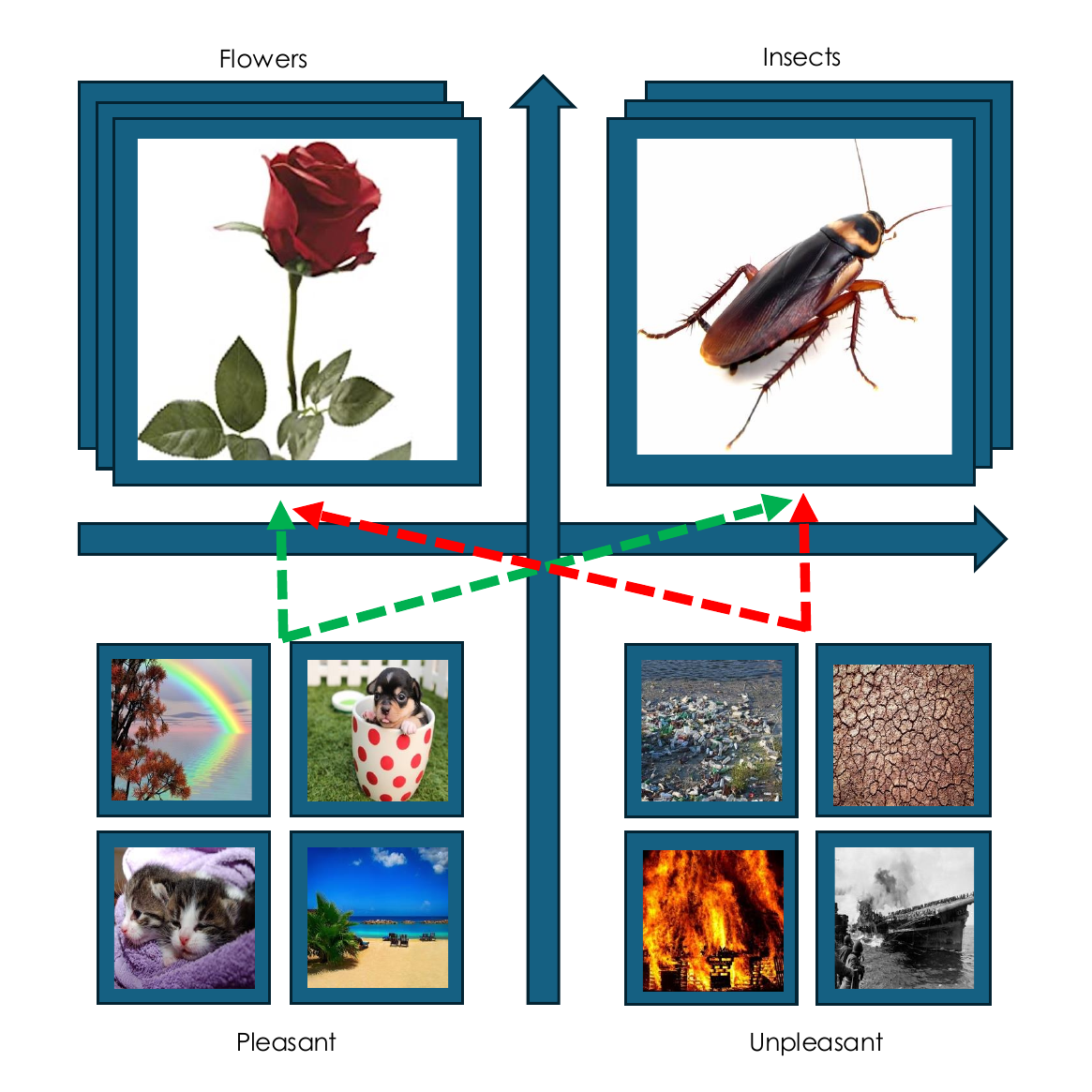}
    \caption{We use Embedding Association Tests, bias evaluation methods for representational or generative models, to quantify biases in 131 CLIP models. Images shown are a subset of the stimuli used to represent the concepts in the Image Embedding Association Test \cite{Steed2021} and our controlled attribute stimuli taken from \citet{kurdi2017introducing}. Distances shown illustrate a stereotype congruent bias (similar to that found in humans), where images representing the concept \textit{Flower} are closer to images representing the concept \textit{Pleasant}, and images representing the concept \textit{Insect} are closer to images representing the concept \textit{Unpleasant}.}
    \label{fig:eat-graphic}
\end{figure}

\section{Background and Related Work}
\label{sec:rw}
We now introduce background information on CLIP models and methods for measuring intrinsic bias in these models.

\noindent\textbf{Unimodal Embedding Association Tests}
\citet{Caliskan2017SemanticsBiases} introduced EATs to measure the associations between static word embeddings which encode concepts related to targets (typically social groups) and attributes, similar to Implicit Association Tests (IATs) for human associations \citep{greenwald1998measuring}. As contextual word embeddings replaced static word embeddings, alternate methods to measure intrinsic bias and associations were also developed \citep{Guo2021DetectingBiases, May2019OnEncoders}. The Sentence Encoder Association Test (SEAT) introduced by \citet{May2019OnEncoders} measures intrinsic bias by operating over a set of  target sentences and attribute sentences which are semantically bleached excluding the words which represent the concepts of interest.  

Additionally, EATs have also been developed for modalities other than text, including vision. \citet{Steed2021} introduce the Image Embedding Association Test (iEAT), which is similar to static word embedding EATs except that it operates over embeddings which represent single images rather than words. EATs in both the textual and visual domains have been shown to replicate biases which are observed in humans \cite{Caliskan2017SemanticsBiases, Steed2021}, making them valuable tools for investigating learned associations in both unimodal and cross-modal models. 

\noindent\textbf{Contrastive Language-Image Pre-training} 
CLIP models are some of the most widely used vision-language models due to their success in zero-shot classification tasks \citep{Radford2021LearningSupervision} as well as their usage as components in popular text-to-image generation systems such as DALL-E and Stable Diffusion. CLIP models have separate image and text encoders, which are connected in a joint cross-modal embedding space \citep{Radford2018ImprovingPre-Training}.  During pre-training, datasets of image-caption pairs comprising hundreds of millions of observations are fed into the models. The model's objective maximizes the cosine similarity between an image and its paired caption, while minimizing the cosine similarity between the image and all other captions in the pre-training batch. Several variations on the original model architecture and training dataset have been proposed to improve CLIP; see Section~\ref{sec:data}.

\noindent\textbf{Biases in Vision Language Models}
\citet{cabello-etal-2023-evaluating} investigate the mechanisms of gender bias amplification in pre-training and fine-tuning stages with vision-language models based on the LXMERT architecture \cite{tan-bansal-2019-lxmert}. This work builds upon theirs by investigating intrinsic biases within models based on the CLIP architecture, which is more commonly used, and through non-human, race, and age bias tests as well as gender bias. Additionally, our method does not rely on having access to training corpora or curating lexicons, making it more flexible to apply to a wider variety of models and biases. \citet{hong2024s} examine biases in CLIP-based pretraining data filtering strategies. Their findings demonstrate that underrepresented populations face higher rates of exclusion than historically well-represented Western demographics.

Most work investigating biases in CLIP models does so only for individual models. For example \citet{Janghorbani2023Multi-ModalModels} study a  CLIP model and find that it tends to associate images representing homosexuality with text such as ``offending'' and ``vulgar,'' while for heterosexual images with words such ``blissful'' and ``awesome,'' among other associations. Further biases in CLIP's embedding space related to race in 3 CLIP models \cite{Wolfe2022MarkednessAI,Wolfe2022EvidenceAI,Wolfe2022AmericanAI}, and gender in 9 CLIP models \cite{Wolfe2023ContrastiveBias}.

The only comparative approach that studies bias in CLIP models to our knowledge \citet{Berg2022ALearning} compares gender bias across 9 CLIP models. They find that larger pre-training datasets tend to lead to decreased bias and hence associated with better zero-shot classification performance, they examine only a small set of biases, model architectures, and training datasets. It is not clear from this work the extent to which these trends would extend to other biases or to other CLIP models.

\section{Experimental Setup }
\label{sec:data}

We now describe the experimental setup and data for upstream pre-training factors, intrinsic bias, and downstream performance.

\noindent\textbf{CLIP Models and Upstream Factors}
The models we study include the original nine CLIP models released by OpenAI \cite{Radford2021LearningSupervision}; 29 models introduced by \citet{Cherti2023ReproducibleLearning}, which were pre-trained on variable-sized English-only subsets of the LAION 5B dataset \cite{Schuhmann2022LAION-5B:Models}; and 93 additional models from the OpenCLIP project \cite{Ilharco2021OpenCLIP}.

The 131 models studied range in size from 102 million to 5 billion parameters, and all use transformer-based text encoders. While most use transformer-based image encoders as well, 17 of the 131 models we study use ResNets or ConvNeXts, convolutional architectures. The pre-training factors that we consider are the size of the model (measured in the number of parameters), model architecture, pre-training dataset, and size of the pretraining dataset (measured in the number of samples). Model architecture has been tied to bias in other modalities \cite{Ladhak2023WhenSummarization}, and model size and pre-training dataset have been tied to bias in nine CLIP models by \citet{Berg2022ALearning}.

The datasets used for pre-training these CLIP models consist of image-caption pairs sourced from the Internet and created under varying levels of supervised curation. For example, the OpenAI WebImageText dataset \cite{Radford2021LearningSupervision} includes pairs whose text contains an element from a set of pre-defined phrases, while the LAION 5B dataset \cite{Schuhmann2022LAION-5B:Models} was filtered to remove images suspected of containing illegal content. Some datasets, such as CC 12M, are revised even further to remove or mentions of social groups or names in order to minimize biases which can be learned from the data. The datasets range in size from 12 million to 5 billion pairs; further statistics for pre-training datasets and architectures are available in the Appendix section \ref{sec:apd_data}.

\noindent \textbf{Embedding Association Test Stimuli}
To measure bias in CLIP models, we use a controlled approach concerning a broad set of concepts across both language and vision modalities. We consider five sets of association tests: non-human (flowers-insects; instruments-weapons), race (European American-African American), gender (women-men), and age (young-old). For all of these target categories, the attribute categories are positively or negatively valenced concepts, due to their strong associations with social groups both in human cognition and unimodal models \citep{toney2020valnorm, harmon2013does}. For each domain, the expected outcome is that the first group is more positively valenced than the second. 

We use the EATs introduced in \citet{Steed2021} and \citet{May2019OnEncoders} for vision (iEAT) and language (SEAT) modalities, respectively, in order to test associations between groups and valence. In SEAT, the human groups are represented both with names and highly associated words in individual tests, meaning there are seven textual EATs and four image EATs. We want to note that the `Gender/Valence' category was not tested in the original SEAT and iEAT studies. We follow previous work \cite{caliskan2022gender, charlesworth2024extracting} that establishes women as being more associated with positive valence compared to men, and thus consider women to be the first group in the gender comparison, to represent the stereotype-congruent direction in our analysis.  Additionally, the `instruments' group was not included in the original iEAT study. Following the text stimuli from \citet{May2019OnEncoders} we carefully curated new image stimuli that satisfy the iEAT requirements. 

Furthermore, we introduce a variation in the iEAT and SEAT attribute stimuli in order to use text and images which are more principled and grounded. Specifically, we use new image stimuli from the OASIS dataset \cite{kurdi2017introducing} and new text stimuli from the NRC-VAD lexicon \cite{mohammad2018obtaining} which contain images and words/phrases respectively that are rated and validated by humans and offer more control and human-grounded valence inputs. Figure \ref{fig:eat-graphic} contains a visualisation of the non-human category EAT using our new stimuli. Further details of these stimuli and how they are selected are provided in the Appendix \ref{apd:eat_data}.

EATs are computed across all modality combinations. iEAT consists only of image targets and image attributes, while SEAT consists only of text targets and text attributes. To perform the cross-modal analysis, we combine image and text stimuli, resulting in additional combinations: images as a target with textual attributes, and text as a target with image attributes. Biases are thus computed for four modality combinations: five \textit{All Image}, eight \textit{All Text}, five \textit{Image as Target}, and eight \textit{Text as Target}, totaling 26 tests.

\noindent\textbf{Downstream Performance (VTAB+)}
Because work from \citet{Berg2022ALearning} found an association between zero-shot performance and bias in nine CLIP models, we also test the relationship between performance and bias for models that have performance data available. We employ performance measured on VTAB+ \cite{Schuhmann2022LAION-5B:Models}, a suite of 35 image classification and retrieval tasks, which includes broad sets of images such as ImageNet \cite{Deng2009Imagenet:Database}, sets of natural images captured with standard or specialized equipment such as Caltech-101 \cite{Li2022Caltech101} or Diabetic Retinopathy \cite{Gulshan2016DevelopmentPhotographs}, as well as structural images, such as SmallNORB \cite{LeCun2004LearningLighting}. 
\section{Approach}
We describe the method we use for quantifying biases in CLIP models, as well as the regression models we use for exploring the relationship between biases, upstream pre-training factors, and downstream performance.

\noindent\textbf{Measuring Intrinsic Bias}
We measure intrinsic bias using EATs \cite{Caliskan2017SemanticsBiases, Guo2021DetectingBiases,Steed2021,Wolfe2022AmericanAI,Wolfe2023ContrastiveBias}, which provide a generalizable and principled method for quantifying biases related to a variety of concepts, such as race and gender, grounded in literature in cognitive and experimental psychology \cite{Blodgett2020LanguageNLP}. An EAT compares similarities between four sets of embeddings created by a model: Two sets of target embeddings which represent social groups, denoted $X$ and $Y$, and two sets of attribute embeddings which represent valence denoted $A$ and $B$ as described in Section \ref{sec:data}. Each EAT gives an effect size $d$, whose magnitude indicates the strength of the bias, calculated as follows:

\[d=\frac{mean_{x \in X}s(x,A,B)-mean_{y \in Y}s(y,A,B)}{std\_dev_{w \in X \cup Y}s(w,A,B)}\]

\noindent where $s$ is given by:

\begin{equation*}
    \begin{split}
        s(w,A,B) =  mean_{a \in A}&{cos(w,a)} \\
        &- mean_{b \in B}{cos(w,b)}\\
    \end{split}
\end{equation*}

\noindent and $cos$ refers to cosine similarity, a distance metric used in quantifying associations by capturing information overlap between embeddings. We order the sets of stimuli such that a positive $d$ value indicates a bias that is congruent with a stereotype that has been documented in society (i.e. \textit{flowers}, \textit{instruments}, \textit{women}, \textit{European American}, and \textit{young} are more associated with \textit{pleasantness}). 

\noindent \textbf{Intrinsic Bias and Upstream Factors} To investigate the relationship between intrinsic bias, measured by the EAT effect size ($d$), and various upstream factors, we employ a mixed effects regression model. The upstream factors considered include the log of parameter size (\textit{log(param)}), model architecture (\textit{arch}), pre-training dataset (\textit{dataset}), and the log of dataset size (\textit{log(dataset size)}). The model is specified as follows:

{ \small \begin{equation*} \begin{split} {d}_{ij} = \beta_0 + \beta_1 \log({param})_{ij} \\ + \beta_2 {arch}_{ij} 
+ \beta_3 {dataset}_{ij} + \beta_4 \log({dataset size})_{ij} \\ + \ u{0j} + u_{1j} \log({param})_{ij} + u{2j} \log({dataset size}){ij} + \epsilon_{ij} \end{split} \end{equation*}}

where $i$ indexes individual observations and $j$ indexes groups defined by modality and test order combinations. Here, $\beta_0$ is the fixed intercept, while $\beta_1$ to $\beta_4$ are fixed coefficients for the predictors. The terms $u_{0j}$, $u_{1j}$ and $u{2j}$ represent random intercepts and slopes for \textit{log(param)} and \textit{log(dataset\_size)}, capturing group-specific baseline $d$ and variability in the effect of model size. The residual error is denoted by $\epsilon_{ij}$.

Significant fixed effects for upstream factors indicate their contribution to intrinsic bias.  For example, a significant positive $\beta_3$ suggests that models trained on certain pre-training datasets exhibit higher intrinsic bias. The inclusion of random effects allows the model to account for unobserved heterogeneity across different groups, thereby enhancing the accuracy and generalizability of the estimates. Reproducibility details are provided in Appendix \ref{apd:mixed_effects}.

\noindent\textbf{Intrinsic Bias and Downstream Performance}
We compute Pearson's correlation between intrinsic bias (EAT effect size $d$) and performance on the VTAB+ benchmark, considering zero-shot classification and captioning tasks relevant to each modality. Correlations are computed separately for each test category and modality combination in order to reveal modality-specific trends in the relationship between intrinsic bias and performance.

\section{Experiments and Results}

Following \citet{May2019OnEncoders} and \citet{Steed2021}, we compute the EAT effect sizes following the SEAT and iEAT methods across the 131 models and 26 tests for a total of 3,406 data points. Our analysis spans four modality combinations: \textit{All Text}, \textit{All Image}, \textit{Image as Target}, and \textit{Text as Target}, across various bias tests including  `Flower-Insect/Valence', `Instrument-Weapon/Valence', `Gender/Valence', `Race/Valence', and `Age/Valence' (as introduced in Section \ref{sec:data}). The EAT effect sizes computed using stimuli from the original SEAT and iEAT studies demonstrate a high overall variance of 0.62. In an effort to offer more control in our experiments thereby making the effect sizes more comparable across models and reducing the impact of outliers, we recompute them across using the newly controlled and grounded attribute stimuli for our tests, drawn from the NRC-VAD lexicon \cite{mohammad2018obtaining} and OASIS \cite{kurdi2017introducing} datasets. 

We observe reduced effect size variance across models and tests to 0.59 overall (a 4.8\% decrease) when replacing SEAT and iEAT attributes with our new attribute stimuli. The decrease in variance was particularly pronounced for the \textit{All Text} modality (a 33.96\% reduction in variance), which suggests that our set is less susceptible to noise and idiosyncrasies that may have plagued previous test sets. Furthermore, changing the stimuli shows better alignment with human stereotypes, showing a significant effect size ($d>=0.2$) in 70.23\% of the 3406 instances using the new stimuli, while this number is lesser at 67.88\% using the old stimuli. 

We note that even after using more controlled stimuli, a high aggregate variance is not surprising, due to the scale and nature of the study. Our experiments in subsequent sections investigate this variation from known dataset choice and model architectural sources. Instead, by employing these carefully curated and grounded stimuli, we gain a clearer lens through which to examine the underlying biases present in various models. Consequently, we present all further results using these new stimuli, providing more robust and generalizable insights into bias in VLMs.

\begin{figure}[h]
    \centering
    \includegraphics[width=\linewidth]{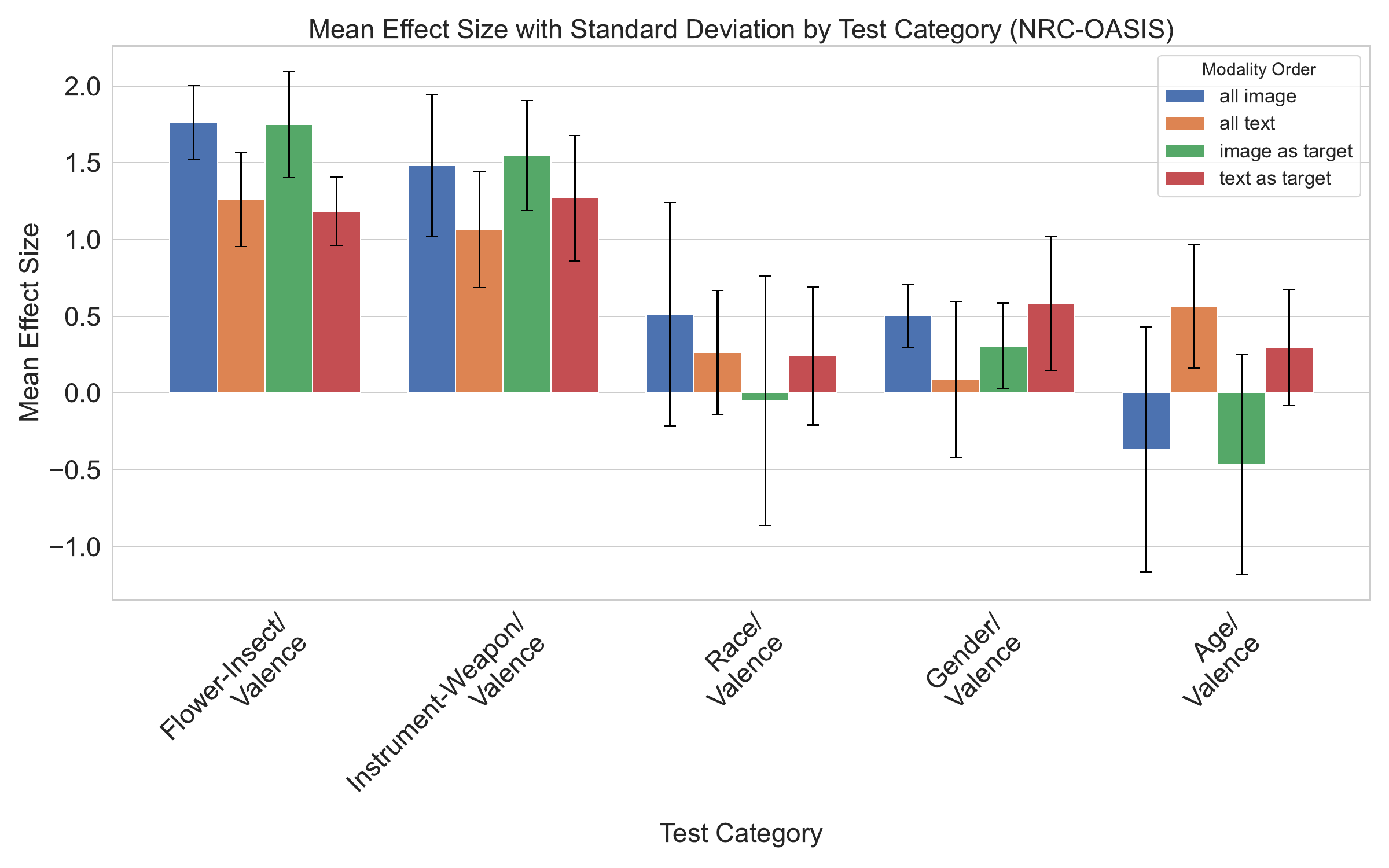}
    \caption{Aggregate Effect Size ($d$) by Test Category Across Modality Orders (NRC-OASIS) along with error bars in black representing standard deviation.}
    \label{fig:eat_aggregate}
\end{figure}

\noindent\textbf{EATs as an Aggregate Measure of Bias} As shown in Figure \ref{fig:eat_aggregate}, 17 out of 20 EATs reveal a pattern of associations which aligns in directionality to results from Implicit Association Tests taken by humans (shown by a positive effect size).\footnote{The magnitude of IAT and EAT effect sizes is not directly comparable because the two tests differ in robustness and contextual dependency. In \ref{apd:iat}, we provide an overview of commonly reported effect sizes in IAT literature for the readers' convenience.} Only in the \textit{All Image} modality for `Age/Valence' and the \textit{Image as Target} for `Race/Valence' and `Age/Valence' are the effect sizes negative, representing associations opposite from those of humans. 

As with humans, `Flower-Insect/Valence,' and `Instrument-Weapon/Valence' show the largest effect sizes across modalities ($d>1$), and associations between valence and social groups are weaker but still present. In all cases, the magnitude of effect sizes varies depending on the modality. Our findings indicate that biases in CLIP models generally align with those found in human assessments in 78.86\% of the 3,406 cases (where $d>0$). In Appendix \ref{apd:old_aggregate}, we show that the direction of effect size across groups is consistent with the original SEAT and iEAT stimuli. 

Figure \ref{fig:eat_aggregate} also contains error bars that represent the standard deviation of effect sizes for the different test categories and modality combinations. We note that there is consistently lower variance for the non-human baselines, such as `Flower-Insect/Valence' and `Instrument-Weapon/Valence', compared to the variance observed in social bias categories like `Age/Valence' and `Race/Valence' indicating that these categories are inherently more susceptible to variability, likely due to the complexity and diversity of social concepts across different training datasets. 

\noindent\textbf{Relationship between Intrinsic Bias and Upstream Factors} We conducted a comprehensive mixed effects regression across the 3,406 observations within 16 different combinations of modality and EAT test category to understand how various upstream factors influence intrinsic bias, measured through the EAT effect size $d$. The model included random slopes and intercepts, effectively capturing high ($\beta=0.46$) group-level variability insights into both fixed and random effects across different combinations of modalities and test categories. For detailed and reproducible information regarding model specifications, variable definitions, and the experimental setup, see Appendix \ref{apd:mixed_effects}.

\begin{figure}[h]
    \centering
    \includegraphics[width=\linewidth]{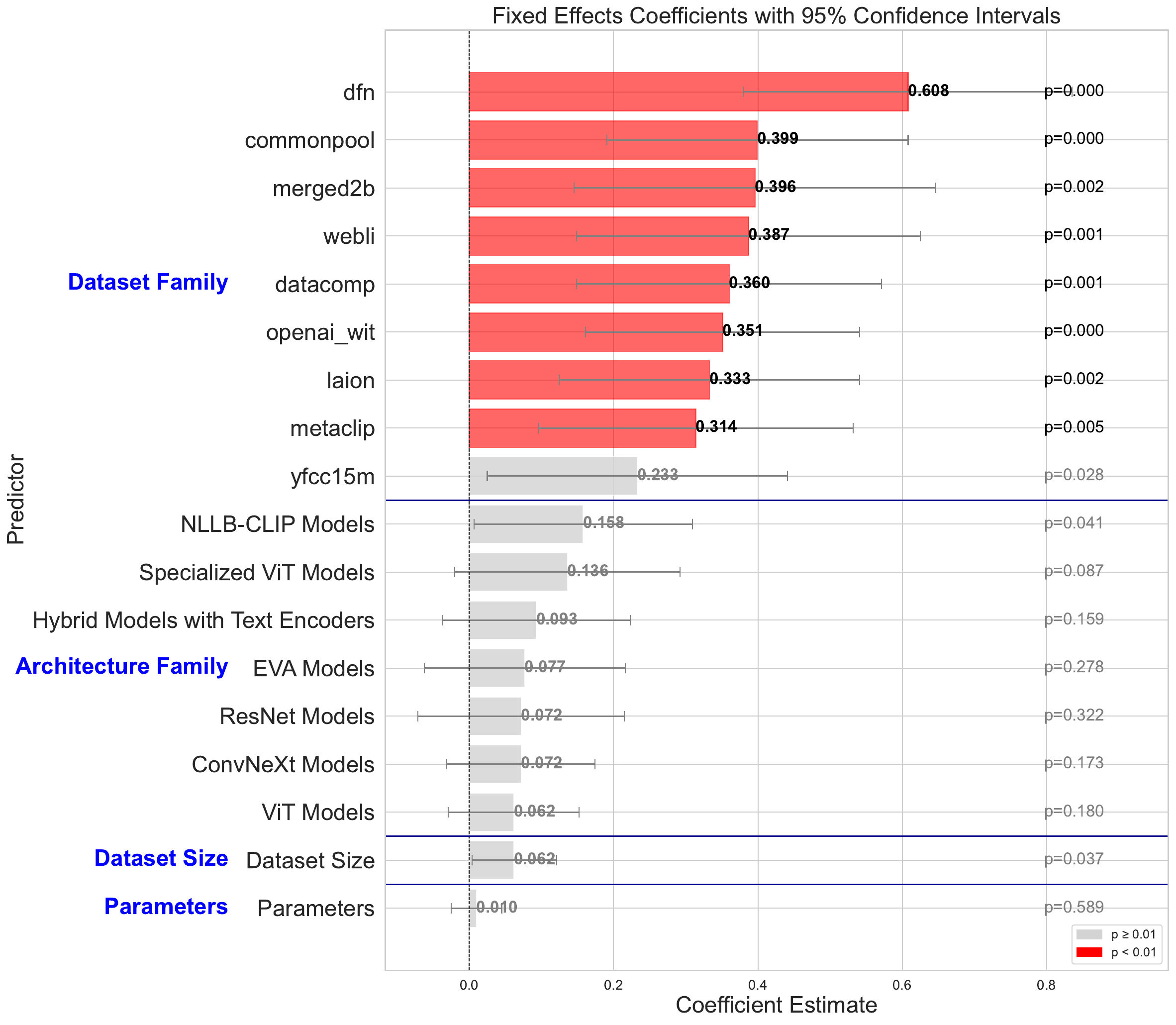}
    \caption{Fixed effects coefficients with 95\% confidence intervals for upstream factors affecting intrinsic bias. The plot illustrates the impact of dataset family, architecture family, dataset size, and model parameters, highlighting statistically significant predictors ($p < 0.01$) in \textbf{red}, while factors that are not significant are greyed.}
    \label{fig:rq2}
\end{figure}

As shown in Figure \ref{fig:rq2}, our findings reveal that dataset family plays a crucial role in determining the magnitude of intrinsic bias. Specifically, several dataset families, including `dfn' ($\beta_{3}=0.608$), `commonpool' ($\beta_{3}=0.399$), `merged2b' ($\beta_{3}=0.396$), `webli' ($\beta_{3}=0.387$), `datacomp' ($\beta_{3}=0.360$), `openai\_wit' ($\beta_{3}=0.351$), `laion' ($\beta_{3}=0.333$), and `metaclip' ($\beta_{3}=0.314$) showed significant positive associations ($p<0.01$) with intrinsic bias effect size with respect to the reference dataset of `CC12m', chosen because we hypothesized its curation strategy would lead to the lowest levels of intrinsic bias. Marginal associations observed in `yfcc15m' and `CC12m' suggests that training on certain datasets contributes more to bias compared to others. This highlights the substantial influence of pretraining data on the biases present in the models.

In contrast, variations in model architecture (although having a positive direction of influence) had no statistically significant impact on intrinsic bias. None of the architectural families demonstrated a significant impact on effect size of bias compared to the reference category, suggesting that, at least within the scope of our study, architectural differences do not play a primary role in influencing bias. Additionally, `log\_params' and `log\_dataset\_size'—did not exhibit significant effects on effect size of bias either.

\noindent\textbf{Relationship between Intrinsic Bias and Downstream Performance}

We investigated the relationship between intrinsic biases measured in CLIP models and their performance on downstream tasks using the VTAB+ benchmark \cite{Schuhmann2022LAION-5B:Models} to understand how intrinsic biases in the models relate to their downstream performance across different modality combinations. Previous research has suggested that biases can influence model performance, particularly as models optimized for accuracy tend to learn and amplify societal biases \cite{hall2022systematic}.

\begin{figure*}[h]
    \centering
    \includegraphics[width=\linewidth]{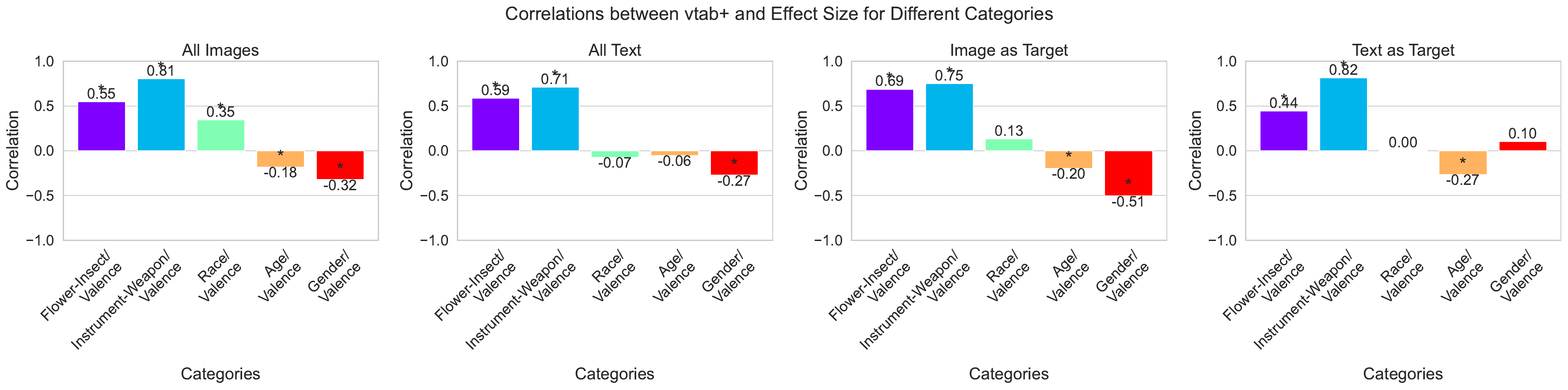}
    \caption{Measure of Pearson's correlation $r$ between effect size magnitude and downstream VTAB+ performance across test categories and modality combinations. Significant values are marked with an asterisk.}
    \label{fig:eat-correlations}
\end{figure*}

Among the different test categories and modality combinations, we observed positive associations between intrinsic bias and downstream performance (meaning increased bias correlates to improved performance) for the non-human categories of `Flower-Insect/Valence' and `Instrument-Weapon/Valence' for   \textit{All Image} ($r=0.55, 0.81$), \textit{All Text} ($r=0.59, 0.71 $), \textit{Image as Target} ($r=0.69, 0.75$), and \textit{Text as Target} ($r=0.44, 0.82$) in addition to the human category `Race/Valence' in the \textit{All Image} combination ($r=0.35$). These are shown in Figure \ref{fig:eat-correlations}. We found negative correlations (meaning increased intrinsic bias correlates to worse performance) for the `Gender/Valence' for \textit{Image as Target} ($r=-0.51$) and \textit{All Text} ($r=-0.27$).  Insignificant correlations were observed primarily for `Race/Valence' and `Age/Valence' in various modalities and `Gender/Valence' in the \textit{Text as Target} modality.

\section{Discussion}
In this work, we explore how intrinsic biases and associations in VLMs are influenced by the interaction between modalities and upstream pretraining factors, and their impact on downstream task performance. Our findings highlight that the magnitude of bias effect sizes depends on the modality combination and test category being observed. However, across the board, the effect sizes are significantly influenced by dataset selection and correlate with model performance on downstream tasks. These insights highlight existing pitfalls in the data and training pipelines of VLMs with respect to fairness considerations and provide important implications for mitigating biases in the future development of these models.

\noindent\textbf{Bias in Cross-Modal Interactions} Our analysis reveals that intrinsic biases manifest differently across combinations of text and image modalities. The `Flower-Insect/Valence' and `Instrument-Weapon/Valence' tests show consistently high effect sizes across all modality combinations, indicating a strong association that is unaffected by modality. `Gender/Valence' biases show positive effect sizes in the \textit{All Image} setting, aligning with previous findings that women are associated with more positive terms than men \cite{caliskan2022gender, charlesworth2024extracting}. 

The representation of bias in models also varies substantially depending on the category of bias and the modality combination. Notably, for `Age/Valence', the direction of the effect sizes differs based on the modality. When analyzing images, older individuals are associated with more positive valence, whereas in text, younger individuals tend to be more positively associated. This discrepancy highlights a critical aspect of modality-specific bias propagation, where the amount and type of information conveyed through visual modalities can differ from that in textual modalities, leading to distinct biases. Such modality-dependent differences in the representation of `Age/Valence' suggest that crossmodal VLMs are influenced by biases specific to each modality. Text stimuli often use older names that are less frequent and therefore may not be represented accurately \cite{wilson2024gender}, while images can convey richer, more nuanced information about age, potentially leading to different bias patterns.

\noindent \textbf{Impacts of Upstream Factors} We focus on identifying which upstream factors—such as dataset characteristics and model design decisions—most significantly influence the intrinsic biases in CLIP models. By examining these biases from both unimodal and crossmodal perspectives, we aim to understand how different training inputs and model architectures contribute to intrinsic bias. This includes an in-depth analysis of how various datasets, including filtering strategies used to curate the selection of datasets, impact the emergence of biases.

In our analysis presented in Figure \ref{fig:rq2}, we demonstrate that the choice of training dataset significantly impacts intrinsic bias, independent of other upstream factors such as model architecture or parameter count. We observe models curated with both automated (e.g. `dfn' \cite{fang_data_2023}) and heuristic filtering strategies (e.g., dataset versions of `commonpool' and `datacomp' \cite{gadre2024datacomp}) to ensure high data quality and subsequent high downstream performance on tasks such as ImageNet accuracy exhibited significantly higher levels of bias, which is likely due to lack of consideration for equitable group identity representation in the dataset curation process. 

Filtering methods that rely on automated neural network-driven decisions \cite{fang_data_2023}, while outperforming heuristic-based approaches in downstream tasks, tend to exacerbate societal biases even further. These results provide strong evidence that bias amplification often originates from the decisions made during the data curation phase, underscoring the need for more ethically-conscious dataset curation practices. 

Our findings align with suggestions from \citet{hong2024s} and \citet{gadre2024datacomp} to exercise caution when using models trained on these datasets to actively make decisions that impact people. 
One potential avenue for dataset-related bias mitigation in CLIP models could be replacing names with a generic "person" token, like in CC 12M \cite{Changpinyo2021ConceptualConcepts} which removed some social group signals contained in the dataset. We hypothesize this may have contributed to lesser bias observed in Figure \ref{fig:rq2}, but the full impact of hypernymization is still unclear and left for future work.

Architecture choice was found to be less impactful compared to dataset selection, which aligns with expectations since most of the text and image encoders in our study were transformer-based, with a few cases being CNN-based image encoders. The synthetic processing in these models was not extensive enough to introduce significant additional bias amplification, and the parameter count remained within a reasonable range without incorporating components, unlike more complex architectures involved in applications like text-to-image generation. 

Dataset size was also not a significant contributor to bias, which contradicts the findings of \citet{Berg2022ALearning}. Our findings indicate that simply increasing the model size or the size of the training dataset does not inherently mitigate or exacerbate intrinsic bias. Instead, other factors such as the composition and characteristics of the dataset are more critical in determining the level of bias.

\noindent \textbf{Effects on Downstream Performance} Our investigation into the correlation between intrinsic biases and downstream task performance, as assessed by VTAB+, reveals significant modal dependencies. We demonstrate that higher intrinsic bias levels correlate with increased performance in downstream tasks across unimodal and crossmodal settings. 

The `Flower-Insect/Valence' and `Instrument-Weapon/Valence' bias shows a high positive correlation across modality combinations suggesting that biases linked to non-human concepts may benefit from consistent training signals, improving model performance and that some associations are universally amplified in conjunction with downstream task performance improvement. For `Gender/Valence' in the \textit{Image as Target} (-0.506 $r$) and  \textit{All Text} (-0.273 $r$) settings, we observed negative correlations, implying that the associations with positive valence increased for the stereotype incongruent `Men' group while model performance improves. This suggests that biases shift as models are further optimized, potentially reinforcing gender-specific stereotypes.

These findings indicate significant modal dependencies in how biases affect downstream task performance. The stark contrast between image-only and text-only settings, particularly test categories that involve social groups such as race and age, suggests that biases are not uniformly propagated across modalities but are instead highly dependent on the type of data and the specific tasks. 

\section{Conclusion}

In this work, we conducted the largest analysis to date on the biases in vision-language models, examining 131 unique CLIP models across 26 datasets and 55 architectures. Our study highlights that the choice of dataset during pre-training, particularly those curated using automatic and heuristic-based filtering approaches that optimise downstream VLM performance, significantly influences intrinsic bias, reinforcing existing disparities. Additionally, we found that biases in models often correlate with improved downstream task performance, across modality settings, suggesting that the possibility that performance optimization can inadvertently amplify certain intrinsic biases as VLMs learn stronger associations between concepts. These findings emphasize the need for more ethically informed dataset curation and bias mitigation strategies to ensure fairer AI models. We release
our code and data at \url{https://github.com/kshitishghate/CLIP_bias}. 
\section{Limitations}

Further empirical studies are needed to compare a broader range of datasets and model configurations to provide a more robust statistical basis for our observations. Our analysis focuses on specific dataset families, model architectures, and parameter sizes. Following the work of \citet{hong2024s} and related work in the text-only domain by \citet{Dodge2021DocumentingLW} and \citet{gururangan-etal-2022-whose}, an in-depth examination of pretraining dataset compositions could offer more insights into mitigating biases effectively. Additionally, there is room for improvement in the measurements of bias using EATs. The stimuli we used are grounded in existing theories, but further controlling the stimuli, such as examining the frequency of stimuli composition \citep{wilson2024gender, wolfe2021low}, could provide a more nuanced understanding of factors that impact bias effect size measures.

Additionally, we only considered monolingual English-based analyses of vision-language models, while with training datasets are curated using multilingual and multicultural sources such as `webli' \cite{chen2022pali} and culture-specific biases present in those sources could also be inherited \cite{ruggeri2023multi}. While our findings are expected to generalize broadly, extending the study to multilingual settings could yield valuable insights. Additionally, focusing primarily on well-established EATs like race, gender, and age leaves out a broader set of possible biases that could be explored in future work, such as those related to socio-economic status or intersectional identities. Limiting the scope of the analysis to these particular biases may risk oversimplifying the complex interrelationships of factors contributing to biased outcomes in VLMs.

\section*{Ethical Considerations}
As vision-language models become increasingly employed in widespread scenarios, the potential for social impact, both positive and negative, grows with it. This study investigates biases within VLMs by explicitly focusing on how these biases are influenced by pre-training factors such as the choice of the training dataset, model architecture and parameter count. We also see how the instrinsic biases directly relate to a number of downstream zero-shot tasks that VLMs are employed for. By doing so, we aim to increase transparency and understanding of how biases are embedded and manifest in the application of VLMs, with the broader goal of promoting the development of fairer AI systems.

The potential applications of our findings include both the improvement and misuse of AI systems. Understanding how intrinsic biases relate to model performance could lead to targeted interventions to reduce bias. However, the same insights could also be used to amplify biases if misapplied. We caution against the use of biased models in high-stakes scenarios such as hiring, healthcare, or law enforcement, where even minor biases can lead to significant ethical consequences. Our intent is to inform researchers, developers, and policymakers of the importance of addressing biases during model development, especially when deploying models in sensitive areas.

To mitigate ethical risks, we advocate for more comprehensive evaluation and auditing frameworks that explicitly quantify and address biases across a diverse set of social categories. This should include incorporating multiple languages and cultural contexts, as well as addressing more diverse and intersectional group identities to ensure the broadest level of inclusivity. Moreover, we believe that transparency in dataset curation and pre-training processes is critical, and encourage the broader research community to prioritize the use of datasets that are both representative and ethically curated.

Lastly, we acknowledge that our own biases as researchers may influence the design and interpretation of our experiments. We strived for impartiality and accuracy, but we recognize that all research inherently carries subjective perspectives. We urge future researchers to build upon our work while expanding its ethical considerations, ensuring a more inclusive and equitable approach to AI development.

\section*{Acknowledgments}
We are grateful to the anonymous reviewers for their helpful feedback. This work was supported by the U.S. National Institute of Standards and Technology (NIST) Grant 60NANB23D194. Any opinions, findings, and conclusions or recommendations expressed in this material are those of the authors and do not necessarily reflect those of NIST.
\bibliography{references}

\appendix

\appendix

\section{Data}
\label{sec:apd_data}
All code and data used in this study will be made available publicly.

\subsection{Datasets Used by Contrastive Language Image Pre-training Models}

Details concerning the datasets that were used for pre-training the CLIP models we study are provided below.

\subsubsection{OpenAI WebImageText}
The OpenAI WIT dataset \cite{Radford2021LearningSupervision} consists of image-caption pairs sourced from the Internet. The dataset only includes pairs whose text included at least one member of a list of common words, bi-grams, and names, taken from Wikipedia and WordNet. Each member in the list of common words, bi-grams, and names was only allowed to be observed in 20,000 image-text pairs, to ensure that they were not over represented. OpenAI WIT contains 400 million pairs, and was not released publicly. We test 9 models trained on the OpenAI WIT dataset. 

\subsubsection{LAION 5 Billion, 2 Billion, 400 Million, 80 Million, and Aesthetics}
The LAION 5B dataset \cite{Schuhmann2022LAION-5B:Models} consists of approximately five billion image-caption pairs. Pairs in the dataset were originally images and alt-text from Common Crawl. After being downloaded, they were passed into the ViT-B/32 CLIP model released by OpenAI (which was trained on OpenAI WIT), and any pairs whose images were not close in cosine distance to their text were filtered out. Images in the dataset that were suspected of containing illegal content were removed, however other potentially harmful images (which make up an estimated 3\% of the dataset) were tagged but kept in the dataset. Of the total pairs, 2.26 billion have English-based captions, while the remaining have captions in other languages or whose language could not be identified. The pairs with English captions make up the LAION 2B dataset, which the LAION 400M and LAION 80M datasets are subsets of. The LAION Aesthetics dataset that was used for pre-training the models we consider consists of approximately 900 million pairs, selected from the LAION 5B dataset for being aesthetically pleasing to human viewers.

\subsubsection{Yahoo-Flickr Creative Commons 15 Million}
The Yahoo-Flickr Creative Commons 15 Million (YFCC15M) dataset \cite{Radford2021LearningSupervision} is a subset of the larger YFCC100M dataset \cite{Thomee2016YFCC100M:Research}. YFCC100M consists of photos and videos uploaded to Flickr between 2004 and 2014, along with titles and descriptions. Around 11 million pairs in the YFCC100M dataset were predicted by a classifier as being related to the concept of "People," (while the most commonly observed concept was "Outdoor," with around 44 million observations), a frequency which we hypothesize may make learning biases related to humans difficult. YFCC15M includes image-text pairs from YFCC100M whose whose title contains natural language, and/or whose description is in English, as many of the text components in YFCC100M seem to consist of generic filenames or descriptions of camera settings  \cite{Radford2021LearningSupervision}. It is unclear the extent to which the distribution of photos in YFCC100M would be reflected in the YFCC15M subset. 

\subsubsection{Conceptual 12 Million}
The Conceptual 12 Million dataset (CC 12M) dataset  \cite{Changpinyo2021ConceptualConcepts} consists of image-text pairs sourced from the Internet. The pairs are filtered based on image format, offensiveness of content, as well as language, capitalization, and other text features. The dataset was hypernymed, whereby names of people were replaced with a special [PERSON] token. Authors of the dataset also state that they examined the dataset for distributional differences between demographic related words, and did not observe any large differences. 

\subsubsection{Commonpool and Datacomp} The Commonpool and Datacomp datasets were constructed as part of the DATACOMP benchmark, aimed at facilitating rigorous research on multimodal dataset design. Commonpool consists of 12.8 billion image-text pairs sourced from Common Crawl, with multiple scales derived by random subsampling, such as Commonpool-XL, -L, and -S. Filtering strategies played a significant role in dataset curation, focusing on removing NSFW content, deduplication, and face blurring to address safety and privacy concerns. DATACOMP benchmarks incorporate subsets like Datacomp-1B, where these filters were applied to create high-quality datasets from the Commonpool index. Content quality was assessed by leveraging metadata like CLIP similarity scores, enabling high zero-shot performance of models trained on these subsets on downstream tasks like ImageNet, often outperforming proprietary datasets like CLIP's WIT \cite{gadre2024datacomp}.

\subsubsection{WebLI} The WebLI dataset was introduced in the development of the PaLI model, with the aim of creating a high-volume, multilingual dataset to train large vision-language models effectively. It consists of 10 billion images with corresponding text in over 100 languages, collected from a range of public web sources. The multilingual nature of WebLI helps test and extend the model’s capabilities across diverse vision and language tasks beyond English-centric training data. This dataset underwent an extensive filtering process to handle noisy data from the internet while allowing a multilingual mix of image-text pairs, which contributed to state-of-the-art results in tasks such as captioning and visual question-answering \cite{chen2022pali}.

\subsubsection{MetaCLIP} MetaCLIP, or Metadata-Curated Language-Image Pre-training, leverages a metadata-driven curation approach to assemble high-quality image-text training datasets. It starts with a raw data pool from Common Crawl and balances the data distribution based on metadata derived from CLIP’s original curation concepts, ensuring a diverse yet informative subset of training pairs. The use of metadata to curate the dataset, rather than solely relying on black-box filtering, enables superior model training outcomes. MetaCLIP has demonstrated competitive performance in zero-shot classification, outperforming CLIP's original WIT dataset on ImageNet when trained on equivalent model architectures, such as ViT-B and ViT-G \cite{xu2023demystifying}.

\subsubsection{Data Filtering Networks (DFN)} The DFN project focused on developing neural networks to filter large-scale, uncurated datasets effectively. The DFN datasets, specifically DFN-2B and DFN-5B, are constructed from Commonpool, using data filtering networks to select high-quality image-text pairs. These networks were optimized for filtering rather than downstream task performance, demonstrating that models trained on filtered datasets could achieve state-of-the-art results on standard benchmarks. For example, DFN-5B enabled a ViT-H model to achieve an 84.4\% zero-shot accuracy on ImageNet, outperforming datasets like DataComp-1B and LAION-2B, thus highlighting the efficacy of neural network-driven filtering over traditional heuristic-based approaches \cite{fang2023data}.

\subsubsection{Merged2B} The Merged2B dataset, used in the EVA-CLIP project, is a carefully curated dataset combining multiple data sources to improve the efficiency and effectiveness of CLIP training. The dataset incorporates two billion image-text pairs from a mix of public sources, filtered to maximize informativeness while minimizing noise. Merged2B is used to train EVA-CLIP models, which include several techniques to reduce computational costs, such as initializing CLIP training with pre-trained EVA representations, and applying optimizations like flash attention. The resulting models, such as EVA-02-CLIP, achieved competitive zero-shot accuracy with significantly fewer training samples compared to models trained on other large datasets like LAION-2B \cite{sun2023eva}.

\subsubsection{Choice of Open-Clip Models}
We consider all pre-trained models available in OpenCLIP (open-clip-torch version 2.16.0) and supported by timm 0.6.12 as of August of 2024, when we performed our measurement. Some models from \citet{Cherti2023ReproducibleLearning} also appear in OpenCLIP, although the models are not labeled as such, so we filter out any models from OpenCLIP that output identical EAT effect sizes with a model from \citet{Cherti2023ReproducibleLearning}. We also remove two models that were trained on identical datasets, used identical architectures, and trained on identical numbers of samples (either measured in batch size or number of epochs, depending on what information was made available) as those given by \citet{Cherti2023ReproducibleLearning}. This leaves us with 131 models from the OpenCLIP repository. The CLIP and OpenCLIP repositories use MIT licenses.

\subsection{Data for measuring EATs}
\label{apd:eat_data}
\subsubsection{SEAT and iEAT tests and stimuli}

\begin{figure*}[h]
    \centering
    \includegraphics[width=\linewidth]{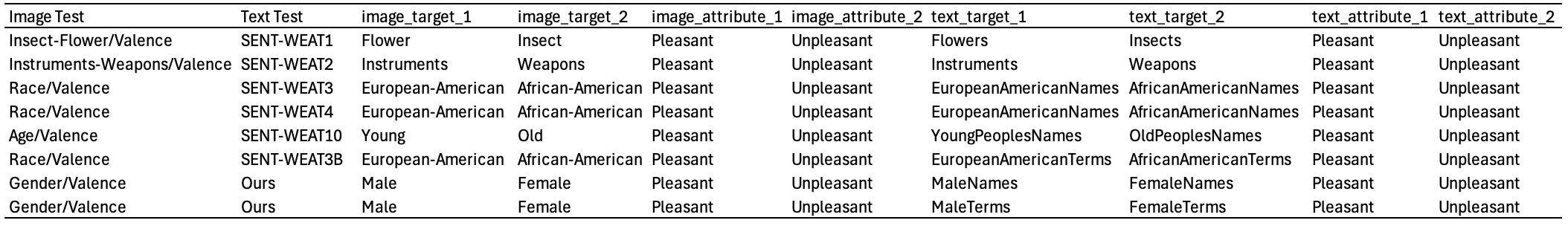}
    \caption{List of Test Categories and Stimuli selected from \citet{May2019OnEncoders} and \citet{Steed2021}.}
    \label{fig:ieat-seat-tests}
\end{figure*}

The specific SEAT and iEAT tests we replicate are presented in Figure \ref{fig:ieat-seat-tests}. For further details necessary for reproducibility, we refer readers to the respective original papers, which provide comprehensive descriptions of these tests and their methodologies.

\subsubsection{Open Affective Standardized Image Set (OASIS) Database}

The Open Affective Standardized Image Set (OASIS) \cite{kurdi2017introducing} contains 900 color images that depict a wide range of themes, including humans, animals, objects, and scenes. These images were rated on valence (positivity or negativity) and arousal (intensity of response) by 822 participants. In our experiments, the 900 images provide non-group-specific valence signals derived from naturalistic image content. All images were standardized to a resolution of 500 x 400 pixels using scaling and cropping processes similar to those described by \citet{wolfe2022evidence}. We selected the top 25 pleasant and top 25 unpleasant images based on valence ratings to construct the two attribute sets used in our analyses. A detailed analysis revealed the aggregate positive valence was equivalent in magnitude to the aggregate negative valence of the filtered images.

\subsection{Textual Templates and Lexica: NRC-VAD Lexicon Valence Stimuli}
Our study employs controlled and balanced textual datasets to investigate how psycholinguistic characteristics of language content influence biases in VLMs. We specifically examine the effects of words and phrases that have psycholinguistic ground-truth ratings in terms of valence. To achieve this, we adopt the sentence template approach used by \citet{May2019OnEncoders} incorporating 6 "semantically bleached templates." These templates are designed to be semantically neutral, providing a consistent syntactic frame for target words without introducing new semantic content. This method ensures that the psycho-semantic attributes of the target words influence the overall sentence meaning. For the experiments, we use the top 25 pleasant and top 25 unpleasant words, sorted by valence ratings, to form the two attribute sets and create the full sets of attributes by using each word in the 6 semantically bleached templates. A detailed analysis revealed the aggregate positive valence was equivalent in magnitude to the aggregate negative valence of the filtered words.

\begin{table*}[htbp]
\centering
\caption{Sentence templates from \citet{May2019OnEncoders}, valence words from \citet{mohammad2018obtaining} and valence images from \citet{kurdi2017introducing} used for the new controlled bias measurement in VLMs.}
\label{tab:templates-and-words}
\begin{tabular}{|l|p{10cm}|}
\hline
\textbf{Type} & \textbf{Content} \\ \hline
Templates & ``This is the word [WORD]", ``That is the word [WORD]", ``There is the word [WORD]", ``Here is the word [WORD]", ``They are the word [WORD]", ``Those are the word [WORD]" \\ \hline
Positive Words & very positive, enjoyable, generous, happily, happy, love, magnificent, extremely positive, sweetie, passionate, cheerful, happier, feelgood, brotherhood, greatness, happiest, joyful, brilliance, smiling, friendliness, joys, laugh, hugs, awesome, superb \\ \hline
Negative Words & shit, nightmare, toxic, horrifying, murderer, homicide, afraid, mistreated, disheartening, angered, bankruptcy, pain, chaos, decayed, murderous, terrorist, cholera, deceit, suffocation, dangerous, shitload, homicidal, hell, genocide, misbehave \\ \hline
Positive Images & Dog 6, Lake 9, Lake 2, Lake 12, Beach 1, Beach 2, Lake 14, Dog 12, Fireworks 2, Rainbow 2, Lake 1, Lake 15, Rainbow 1, Cat 5, Penguins 2, Lake 8, Dog 4, Siblings 1, Dog 18, Baby 1, Lake 13, Fireworks 1, Lake 10, Baby 5, Sunset 3 \\ \hline
Negative Images & Destruction 4, Explosion 5, Scary face 1, War 1, Fire 11, Fire 7, Fire 5, War 8, Severed finger 1, Garbage dump 4, Animal carcass 5, Dirt 1, Garbage dump 2, Fire 9, Tumor 1, Injury 4, War 6, KKK rally 1, Dead bodies 3, Dog 26, KKK rally 2, Dead bodies 2, Dead bodies 1, Dummy 1, Miserable pose 3 \\ \hline
\end{tabular}
\end{table*}

\section{Estimation of Mixed Effects Models}
\label{apd:mixed_effects}

The mixed effects model was chosen to explore the relationship between intrinsic bias (EAT effect size) and upstream factors while accounting for group-level variability across modality and test orders. Initial attempts to fit the model with individual dataset and architecture categories led to convergence issues due to high parameter complexity. To address this, we simplified the analysis by grouping datasets and architectures into broader families, resulting in a more stable and interpretable model. 

The final model included fixed effects for log-transformed parameters ($\log(\text{param})$), architecture family, dataset family, and log-transformed dataset size ($\log(\text{dataset size})$). Random effects were specified for group-level intercepts and slopes for $\log(\text{param})$ and $\log(\text{dataset size})$, capturing differences across modality and test order groups. Including random slopes significantly improved the model's log-likelihood, demonstrating the importance of accounting for the varying influence of parameters and dataset size across groups.

\begin{figure*}[h]
    \centering
    \includegraphics[width=0.8\linewidth]{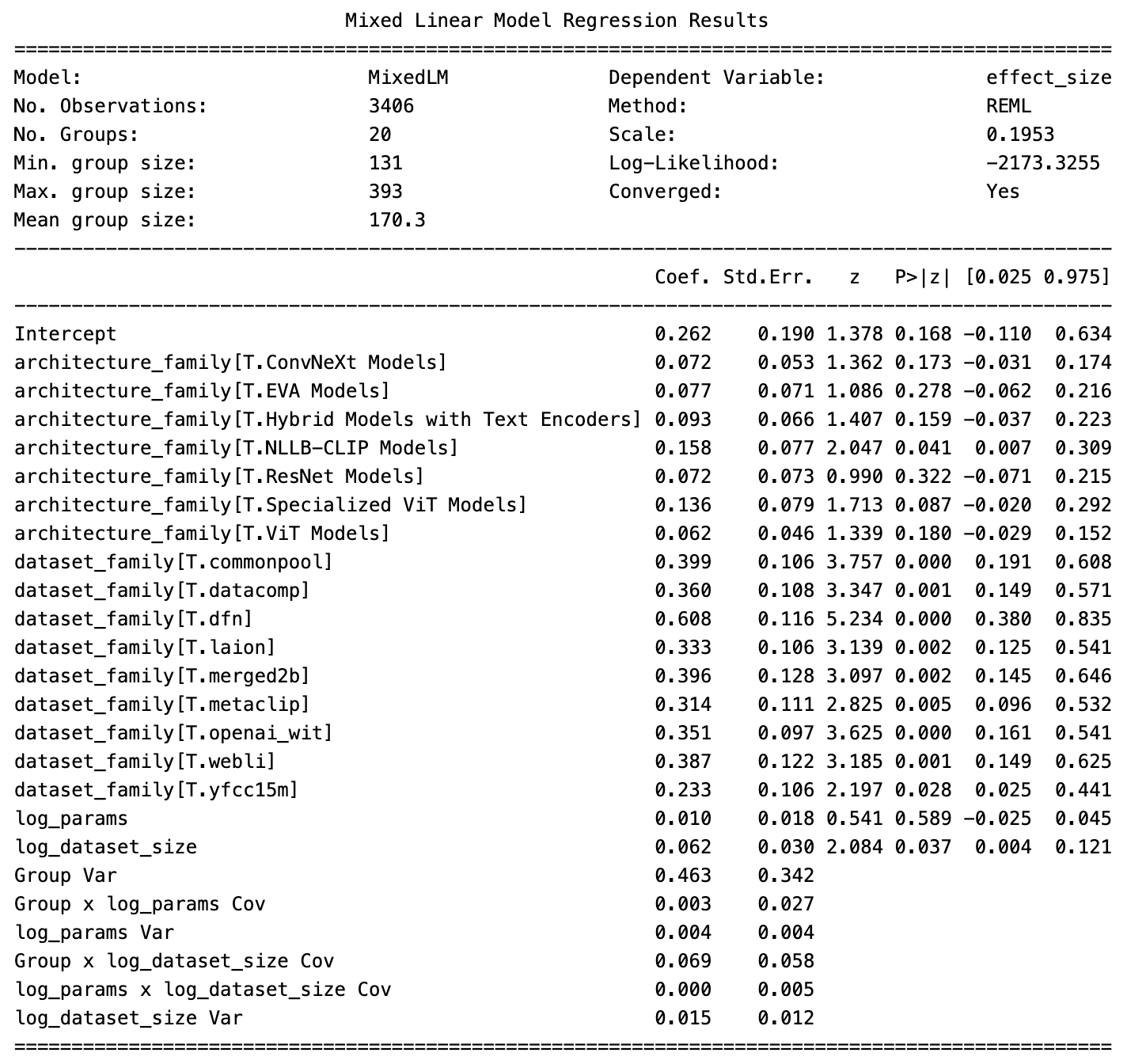}
    \caption{Full mixed effects regression results to measure impact of upstream factors on intrinsic bias.}
    \label{fig:mixed_effects_model}
\end{figure*}

Preprocessing involved removing missing values, log-transforming continuous variables, and scaling using `StandardScaler' to improve numerical stability. Categorical variables were appropriately converted to categorical types for the model. The model was fitted using Restricted Maximum Likelihood (REML) with the `lbfgs' optimizer, achieving a log-likelihood of approximately -2173.32, indicating that the inclusion of random intercepts, slopes and a simplified dataset analysis contributed to better model fit and generalizability. Statistical significance was computed by comparing the Wald t-values to the z-distribution \citep{luke2017evaluating}. Figure \ref{fig:mixed_effects_model} contains the full results of the mixed effects model.

\section{Ablation Results}
\subsubsection{Baseline Results for Old Stimuli}
\label{apd:old_aggregate}

\begin{figure}[h]
    \centering
    \includegraphics[width=\linewidth]{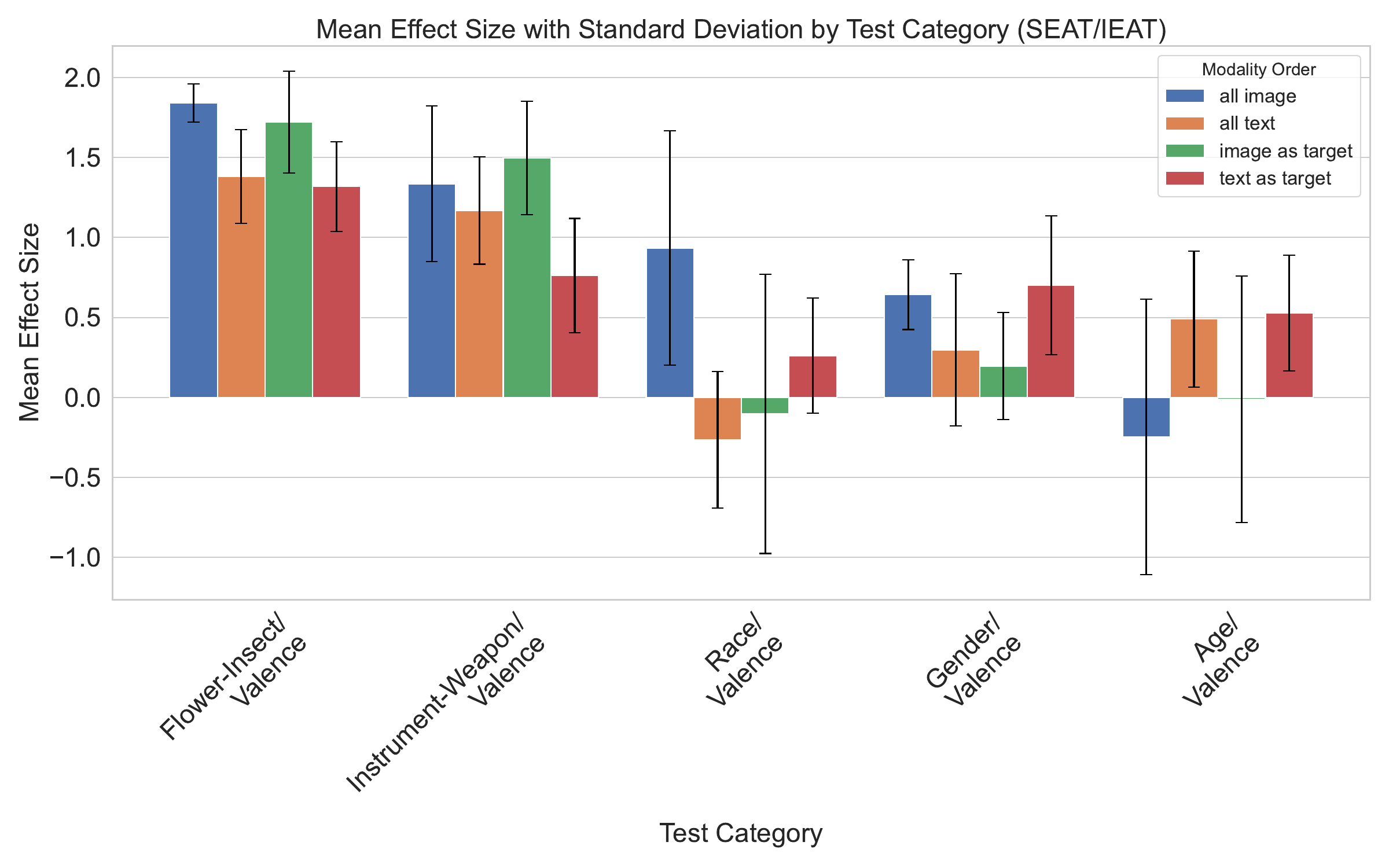}
    \caption{Aggregate Effect Size ($d$) with error bars representing standard deviation by Test Category Across Modality Orders from SEAT and iEAT stimuli. The direction of effect sizes are largely consistent with the effect sizes obtained from new stimuli in Figure \ref{fig:eat_aggregate}. }
    \label{fig:old_effect_size}
\end{figure}

\section{Implicit Association Test Effect Sizes}
\label{apd:iat}

\begin{table*}[]
\centering
\begin{tabular}{|l|l|l|}
\hline
\textbf{Association Test} & \textbf{Mean IAT Effect Size}         & \textbf{\begin{tabular}[c]{@{}l@{}}EAT effect size \\ ranges across \\ modality \\ combinations\end{tabular}}
\\ \hline
Insect-Flower/Valence     & \begin{tabular}[c]{@{}l@{}}1.35 \\ \citep{Greenwald1998}\end{tabular}     & 1.341 ± 0.446 
\\ \hline
Instrument-Weapon/Valence     & \begin{tabular}[c]{@{}l@{}}1.66 \\ \citep{Greenwald1998}\end{tabular}     & 1.490 ± 0.390                                                                                                                                                                                                 \\ \hline
Race/Valence              & \begin{tabular}[c]{@{}l@{}}1.17 \\ \citep{Greenwald1998}\end{tabular}     & 0.248 ± 0.552                                                                                                 \\ \hline
Gender/Valence            & \begin{tabular}[c]{@{}l@{}}(0.02, 0.44) \\ \citep{baron2013}\end{tabular} & 0.361 ± 0.463                                                                                                \\ \hline
Age/Valence               & \begin{tabular}[c]{@{}l@{}}1.42 \\ \citep{Greenwald1998}
\end{tabular}     & 0.007 ± 0.743                                                                                                \\ \hline
\end{tabular}
\caption{Comparison of commonly reported effect sizes found when performing implicit association tests with human subjects vs. the mean and standard deviation of effect sizes observed across modalities for EATs targeting associations between the same concepts.}
\end{table*}

\end{document}